# Unsupervised-learning-based method for chest MRI–CT transformation using structure constrained unsupervised generative attention networks

Hidetoshi Matsuo, M.D.[1,*], Mizuho Nishio, M.D., Ph.D.[1], Munenobu Nogami, M.D., Ph.D., FANMB[1], Feibi Zeng, M.D., Ph.D.[1], Takako Kurimoto, Ph.D.[2], Sandeep Kaushik, MSc.[3], Florian Wiesinger, Ph.D.[3], Atsushi K. Kono, M.D., Ph.D.[1], and Takamichi Murakami, M.D., Ph.D.[1]

[*] Corresponding author
[1] Department of Radiology, Kobe University Graduate School of Medicine, Kobe, Japan
[2] GE Healthcare, Hino, Japan
[3] GE Healthcare, Munich, Germany

**Abstract**

The integrated positron emission tomography/magnetic resonance imaging (PET/MRI) scanner simultaneously acquires metabolic information via PET and morphological information using MRI. However, attenuation correction, which is necessary for quantitative PET evaluation, is difficult as it requires the generation of attenuation-correction maps from MRI, which has no direct relationship with the gamma-ray attenuation information. MRI-based bone tissue segmentation is potentially available for attenuation correction in relatively rigid and fixed organs such as the head and pelvis regions. However, this is challenging for the chest region because of respiratory and cardiac motions in the chest, its anatomically complicated structure, and the thin bone cortex. We propose a new method using unsupervised generative attentional networks with adaptive layer-instance normalisation for image-to-image translation (U-GAT-IT), which specialised in unpaired image transformation based on attention maps for image transformation. We added the modality-independent neighbourhood descriptor (MIND) to the loss of U-GAT-IT to guarantee anatomical consistency in the image transformation between different domains. Our proposed method obtained a synthesised computed tomography of the chest. Experimental results showed that our method outperforms current approaches. The study findings suggest the possibility of synthesising clinically acceptable computed tomography images from chest MRI with minimal changes in anatomical structures without human annotation.

**Abbreviations**

Deep convolutional neural networks (DCNNs)
Generative adversarial networks (GANs)
Zero echo time (ZTE)
Modality-independent neighbourhood descriptor (MIND)
Positron emission tomography/magnetic resonance imaging (PET/MRI)

**1.Introduction**

1.1 Background

New methods of machine learning, such as deep convolutional neural networks (DCNNs), have been recently developed because of easy access to large datasets and computational resources, and DCNN has made remarkable progress in various fields. The performance of DCNNs has significantly improved in the field of image recognition research. Generative adversarial networks (GANs) have received considerable attention in neural networks. For example, Zhu et al. [1] developed an unsupervised learning method that enables the transformation of images between two types of domains using GANs called CycleGAN [2]. They showed that it is possible to transform images between horse and zebra and between day and night.

An integrated positron emission tomography/magnetic resonance imaging (PET/MRI) scanner is the only modality that can obtain metabolic information with PET and morphological information with high soft-tissue contrast using MRI by simultaneous acquisition. Although the advantage of PET/MRI is the accuracy of the fusion images, a major drawback of PET/MRI is the difficulty in attenuation correction for PET reconstruction, which is necessary for the quantitative evaluation of PET. X-ray-based attenuation correction, which is a method of translating CT images from the effective X-ray energy to attenuation coefficients at the PET energy (511 keV), is widely employed for attenuation correction of PET/CT. However, the generation of attenuation-correction maps from MRI (a synthesised CT) is necessary for PET/MRI because no direct relationship exists between gamma-ray attenuation information and MRIs. Moreover, only four-tissue segmentations (air, lung, fat, and soft tissue) other than bone are used for synthesised CT generation because of the difficulties in extracting signals from tissues with low proton density, such as bone tissue, on conventional MR sequences [3].

The zero echo time (ZTE) MR sequence enables imaging of tissues with short T2 relaxation time and is utilised for bone and lung imaging [4–9]. For the head region, ZTE-based attenuation correction is already available in commercial PET/MRI scanners because delineation and segmentation of bone tissue on simultaneously acquired ZTE is relatively easy for rigid and fixed organs [10–13]. A deep learning approach based on the use of paired training data for generating synthesised CT from MRI is now applicable to the head and pelvis regions [14–16]. In the chest, however, bone segmentation from ZTE remains difficult to perform for accurately synthesised CT generation due to its respiratory and cardiac motion, anatomically complicated structure, and relatively thin cortex of the bone.

CycleGAN has been successfully used for medical image analysis, such as cone-beam-CT-CT conversion [17] and MRI–CT conversion of the head [18]. In addition to CycleGAN, other unsupervised learning methods for interdomain image transformations have been proposed and used in medical image analysis [19]. Image transformation using GANs faces a problem in that anatomical consistency cannot be guaranteed.

In the current study, to generate synthesised CT from the ZTE of PET/MRI, we utilised a new unsupervised method called Unsupervised Generative Attentional Networks with Adaptive Layer-Instance Normalisation for Image-to-Image Translation (U-GAT-IT), which is specialised in unpaired image transformation based on attention maps for image transformation [20]. To guarantee anatomical consistency in the image transformation between different domains (synthesised CT and ZTE), the modality-independent neighbourhood descriptor (MIND) [21] was added to the loss of U-GAT-IT. Using our proposed method, the ZTE of PET/MRI of the chest can be converted to synthesised CT. The U-GAT-IT and CycleGAN models were not originally developed for use in medical image analysis. The anatomical structure might differ significantly in the ZTE and synthesised CT images obtained using the U-GAT-IT or CycleGAN methods.

However, to use the synthesised CT as an attenuation correction map for PET/MRI, differences in anatomical structures such as bone, body, and upper arm contour are critical. Our proposed method with U-GAT-IT and MIND successfully prevented anatomical inconsistencies between ZTE and synthesised CT.

1.2 Related work

Transformation in medical images is required in numerous clinical fields, and several applications have been reported, such as noise reduction, MRI–CT transformation, and segmentation tasks. In medical images, however, assembling numerous labelled images for training is challenging. In addition, obtaining an exactly aligned pair of images for inter-modality transformation is difficult. Paired training data for the head and pelvis can be prepared by matching the shapes of CT and MRI using nonlinear image registration, whereas prepared such data for the chest is difficult. Although there have been reports on MRI–CT transformation in the head and pelvis, which are relatively unchanged by body posture [22, 23], it is difficult to obtain an aligned pair of corresponding images of the chest and other regions because of breathing and differences in body posture between MRI and CT. Thus, inter-modality image conversion in the chest has been considered challenging to accomplish.

CycleGAN, which enables unpaired image conversion without the need for directly corresponding images, has attracted attention. It has performed well in various fields, such as the generation of synthesised CT from cone-beam CT [24], CT segmentation [25], and X-ray angiography image generation [26]. Generally, CycleGAN is employed to perform transformations between two types of image domains. However, no direct constraint exists on the structure of the input and output images, and the structural alignment between the input and output images is not guaranteed. In medical images, the transforms of the anatomical structures are critical. To overcome this problem, several studies have been performed; CycleGAN has been extended to three-dimensional medical images [27–29], and loss of CycleGAN was changed to set constraints on anatomical structural change [28, 30, 31].  In addition, various approaches have been attempted such as a deformation-invariant CycleGAN (DicycleGAN) [30], an extension of CycleGAN by adding the gradient consistency loss to improve the accuracy at the boundaries [32], and the use of CycleGAN for the paired data [33].

1.3 Contributions

The contributions of this study are summarised as follows. First, this paper presents a method for performing chest MRI–CT (ZTE to synthesised CT) transformation using unsupervised learning methods such as U-GAT-IT and CycleGAN, which enable unpaired image transformations. Second, the proposed allows constraints to be applied to U-GAT-IT and CycleGAN to overcome the effect of changes in anatomical structures when transforming chest MRI–CT images. For this purpose, we added MIND to the losses of U-GAT-IT and CycleGAN and attempted to suppress the irregular changes in anatomical structures. Third, the combined use of U-GAT-IT and MIND made it possible to generate clinically acceptable synthesised CT images with less structural changes compared with CycleGAN with and without MIND. Fourth, without using any human annotations, the unsupervised learning methods of U-GAT-IT and CycleGAN allowed us to generate synthesised CT.

The remainder of this paper proceeds as follows. Section 2 describes the details of Cycle GAN and U-GAT-IT and the loss of these networks using MIND. Section 3 describes the details of PET/MRI and CT imaging and experimental studies used for the performance comparison. Section 4 describes the experimental results. Section 5 discusses the results and compares them with previous studies; finally, the study is concluded in Section 6.

**2. Materials and methods**

In this study, CycleGAN and U-GAT-IT were used to perform MRI–CT conversion using unpaired data. In addition, we applied MIND, which was proposed in a previous study, to these networks to prevent misalignment between MRI and synthesised CT images. Please refer to Figure 1 for an outline of the proposed U-GAT-IT+MIND process.

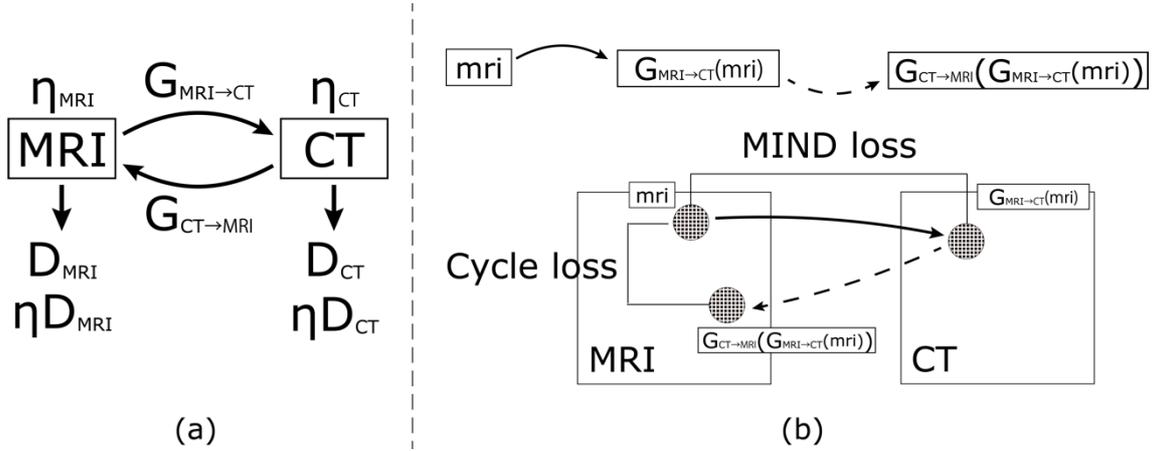

**Figure 1** Outline of proposed U-GAT-IT + MIND process (G, D, and η denote generator, discriminator, and auxiliary classifier, respectively). We introduce Cycle loss, which is a comparison within the same domain after two rounds of transformation; this is in addition to MIND loss, which is a comparison between different domains after one round of transformation.

To compare the performance of U-GAT-IT + MIND loss, we evaluated the performance of CycleGAN alone, U-GAT-IT alone, and the CycleGAN + MIND loss.

2.1 CycleGAN

CycleGAN, developed in 2016, is a method that allows transformations between two different image domains. CycleGAN involves competing networks of an image generator (generator) and an adversarial network (discriminator) that attempt to distinguish the generated synthetic image from the real image. Taking the transformation between MRI and CT images as an example, there is a loss (G loss) to make the synthesised CT image closer to the real CT image for the generator, and a loss (D loss) to distinguish the synthesised CT image from the real CT image for the discriminator. In addition, there are two types of losses in CycleGAN: cycle loss and identity loss. Cycle loss is the difference between the original image and the double-synthesised MRI, which is further synthesised from the synthesised CT based on MRI. Identity loss is the difference between the output image and the input image (CT image and synthesised CT image) when the CT image is input to the CT generator. The same four types of losses are calculated for CT-MRI conversion (when synthesised MRI is generated from real CT). Please refer to the original paper on the conceptual diagram. The three types of losses are as follows — Equations (1)–(4).

***Generator and discriminator loss*** *(Generator and discriminator losses are employed to match the distribution of the translated images to the distribution of the target image):*

$$L_{GAN}(G_{MRI \rightarrow CT}, D_{CT}, I_{MRI}, I_{CT}) = \mathbb{E}_{ct \sim p_{data}(I_{CT})}[\log D_{CT}(ct)] + \mathbb{E}_{mri \sim p_{(data)(I_{MRI})}}[\log(1 - D_{CT}(G_{MRI \rightarrow CT}(mri)))] \quad (1)$$

***Cycle loss*** *(To alleviate the mode collapse problem, we applied a cycle consistency constraint to the generator):*

$$L_{cyc}(G_{MRI \rightarrow CT}, G_{CT \rightarrow MRI}, I_{MRI}, I_{CT}) = \mathbb{E}_{mri \sim p_{(data)(I_{MRI})}}[\| G_{CT \rightarrow MRI}(G_{MRI \rightarrow CT}(mri)) - mri \|_1]$$
$$+ \mathbb{E}_{ct \sim p_{(data)(I_{CT})}}[\| G_{MRI \rightarrow CT}(G_{CT \rightarrow MRI}(ct)) - ct \|_1] \quad (2)$$

***Identity loss*** *(To ensure that the distributions of input image and output image are similar, we applied an identity consistency constraint to the generator):*

$$L_{identitiy}(G_{MRI \rightarrow CT}, G_{CT \rightarrow MRI}, I_{MRI}, I_{CT}) = \mathbb{E}_{ct \sim p_{(data)(I_{CT})}}[\| G_{MRI \rightarrow CT}(ct) - ct \|_1]$$
$$+ \mathbb{E}_{mri \sim p_{(data)(I_{MRI})}}[\| G_{CT \rightarrow MRI}(mri) - mri \|_1] \quad (3)$$

***Sum of losses*** *(Finally, we jointly trained the generators, and discriminators to optimize the final objective):*

$$L_{CycleGAN}(G_{MRI \rightarrow CT}, G_{CT \rightarrow MRI}, D_{MRI}, D_{CT}, I_{MRI}, I_{CT}) = L_{GAN}(G_{MRI \rightarrow CT}, D_{CT}, I_{MRI}, I_{CT})$$
$$+ L_{GAN}(G_{CT \rightarrow MRI}, D_{MRI}, I_{CT}, I_{MRI}) + \lambda_1 L_{cyc}(G_{MRI \rightarrow CT}, G_{CT \rightarrow MRI}, I_{MRI}, I_{CT})$$
$$+ \lambda_2 L_{identity}(G_{MRI \rightarrow CT}, G_{CT \rightarrow MRI}, I_{MRI}, I_{CT}) \quad (4)$$

where $I_{CT}$ denotes the CT image, $I_{MRI}$ denotes the MRI image, $G_{CT \rightarrow MRI}$ denotes the generator that generates MRI from CT, $G_{MRI \rightarrow CT}$ denotes the generator that generates CT from MRI, $D_{CT}$ denotes the discriminator that discriminates between $G_{MRI \rightarrow CT}(mri)$ and $ct$, $D_{MRI}$ denotes the discriminator that discriminates between $G_{CT \rightarrow MRI}(ct)$ and $mri$, and $L_{GAN}$ denotes the loss that includes G loss and D loss. $L_{cyc}$ is the cycle loss, and $L_{identitiy}$ is the identity loss. $\lambda_1$ and $\lambda_2$ denote coefficients of losses.

Finally, the model was trained by reducing the losses using Equation (5):

$$\arg \min_{G_{MRI \to CT}, G_{CT \to MRI}} \max_{D_{MRI}, D_{CT}} L_{CycleGAN} \quad (5)$$

## 2.2 U-GAT-IT

U-GAT-IT is an unsupervised generative attentional network with adaptive layer-instance normalisation for image-to-image translation, which was developed in 2019. Similar to CycleGAN, U-GAT-IT uses the encoder–decoder method for image generation but incorporates the attention module in the discriminator and generator and combines them with the adaptive layer-instance normalisation function (AdaLIN) to focus on the more important parts of the image.

***Generator loss and discriminator loss*** *(Generator and discriminator losses are employed to match the distribution of the translated images to the distribution of the target images.)*：

$$L'_{GAN}(G_{MRI \to CT}, D_{CT}, I_{MRI}, I_{CT}) = \mathbb{E}_{ct \sim p_{data}(I_{CT})}[(D_{CT}(ct)^2] \\ + \mathbb{E}_{mri \sim p_{data}(I_{MRI})}[(1 - D_{CT}(G_{MRI \to CT}(mri)))^2] \quad (6)$$

CAM loss represents the loss that is important for the conversion from MRI and CT based on the information of auxiliary classifiers $\eta_{MRI}$ and $\eta_{CT}$.

***CAM losses*** *(By exploiting the information from the auxiliary classifiers $\eta_{CT}$, $\eta_{MRI}$, $\eta D_{CT}$, and $\eta D_{MRI}$, given an image from $I_{CT}$ $I_{MRI}$. $G_{MRI \to CT}$ and $D_{CT}$ identify where they need to improve in terms of what makes the most difference between two domains.)* :

$$L_{cam}^{G_{MRI \to CT}}(G_{MRI \to CT}, I_{MRI}, I_{CT}, \eta_{MRI}, \eta_{CT}) = -(\mathbb{E}_{mri \sim p_{(data)(I_{MRI})}}[\log(\eta_{MRI}(mri))] \\ + \mathbb{E}_{ct \sim p_{(data)(I_{CT})}}[\log(1 - \eta_{CT}(ct))]) \quad (7)$$

$$L_{cam}^{D_{CT}}(G_{MRI \to CT}, I_{MRI}, I_{CT}, \eta D_{MRI}, \eta D_{CT}) \\ = \mathbb{E}_{ct \sim p_{(data)(I_{CT})}}\left[(\eta D_{CT}(ct))^2\right] + \mathbb{E}_{mri \sim p_{(data)(I_{MRI})}}[(1 - \eta D_{CT}(G_{MRI \to CT}(mri)))^2] (8)$$

*Sum of losses Finally, we jointly trained the encoders, decoders, discriminators, and auxiliary classifiers to optimize the final objective :*

$$\begin{aligned}
L_{U-GAT-IT}&(G_{MRI\to CT}, G_{CT\to MRI}, D_{MRI}, D_{CT}, I_{MRI}, I_{CT}, \eta_{MRI}, \eta_{CT}, \eta D_{MRI}, \eta D_{CT}) \\
&= L'_{GAN}(G_{MRI\to CT}, D_{CT}, I_{MRI}, I_{CT}) + L'_{GAN}(G_{CT\to MRI}, D_{MRI}, I_{CT}, I_{MRI}) \\
&+ \lambda_1 L_{cyc}(G_{MRI\to CT}, G_{CT\to MRI}, I_{MRI}, I_{CT}) + \lambda_2 L_{identity}(G_{MRI\to CT}, G_{CT\to MRI}, I_{MRI}, I_{CT}) \\
&+ \lambda_3 L_{cam}^{G_{MRI\to CT}}(G_{MRI\to CT}, G_{CT\to MRI}, I_{MRI}, I_{CT}, \eta_{MRI}, \eta_{CT}) + \lambda_3 L_{cam}^{G_{CT\to MRI}}(G_{CT\to MRI}, G_{MRI\to CT}, I_{CT}, I_{MRI}, \eta_{CT}, \eta_{MRI}) \\
&+ \lambda_3 L_{cam}^{D_{CT}}(G_{MRI\to CT}, I_{MRI}, I_{CT}, \eta D_{MRI}, \eta D_{CT}) \\
&+ \lambda_3 L_{cam}^{D_{MRI}}(G_{CT\to MRI}, I_{CT}, I_{MRI}, \eta D_{CT}, \eta D_{MRI})
\end{aligned}$$

where $L'_{GAN}$ denotes the loss that includes G loss and D loss. $L_{cam}^{G_{MRI\to CT}}$ is the CAM loss of $G_{MRI\to CT}$, $L_{cam}^{D_{CT}}$ is the CAM loss of $D_{CT}$, $L_{cyc}$ is the cycle loss, and $L_{identitiy}$ is the identity loss. $\lambda_1, \lambda_2$, and $\lambda_3$ denote coefficients of losses. Finally, the model was trained by reducing the losses using Equation (10):

$$\arg \min_{G_{MRI\to CT}, G_{CT\to MRI}, \eta_{MRI}, \eta_{CT}} \max_{D_{MRI}, D_{CT}, \eta D_{MRI}, \eta D_{CT}} L_{U-GAT-IT} \qquad (10)$$

2.3 MIND

MIND is a modality-independent neighbourhood descriptor for multi-modal deformable registration reported by Heinrich et al. in 2012 [21]. MIND can extract numerical descriptors preserved across modalities by extracting local feature structures.

$$I_{MIND}(I, x, r) = \frac{1}{n}\exp\left(-\frac{D_p(I, x, x+r)}{V(I, x)}\right) r \epsilon R, \qquad (11)$$

where *I* denotes the image, *n* denotes the normalisation constant (so that the maximum value equals 1), and $r \epsilon R$ defines the region to be calculated; $D_p(I, x, x+r)$ denotes the distance metric between the positions $x$ and $x + r$; it is expressed by Equation (12). In this study, we considered *r* = 9. Calculations were performed by convolution, as in previous studies [31]. *P* represents a collection of quantities that shifts the image. In this case, there exist 81 sets that shift the image from -4 to 4 along the X- and Y-axis directions.

$$D_p(I, x, x+r) = \sum_{p\epsilon P}\bigl(I(x+p) - I(x+r+p)\bigr)^2 \qquad (12)$$

The denominator $V(I, x)$ represents an estimation of the local variance, and it can be expressed as

$$V(I, x) = \sum_{n \epsilon N} D_p(I, x, x + n), \tag{13}$$

where *N* denotes the 3-neighbourhood of voxel *x*. The left image in Figure 2 presents the original ZTE MRI while the corresponding right image presents MIND. The outlines of the body surface, lungs, bones, and blood vessels in the lungs were extracted.

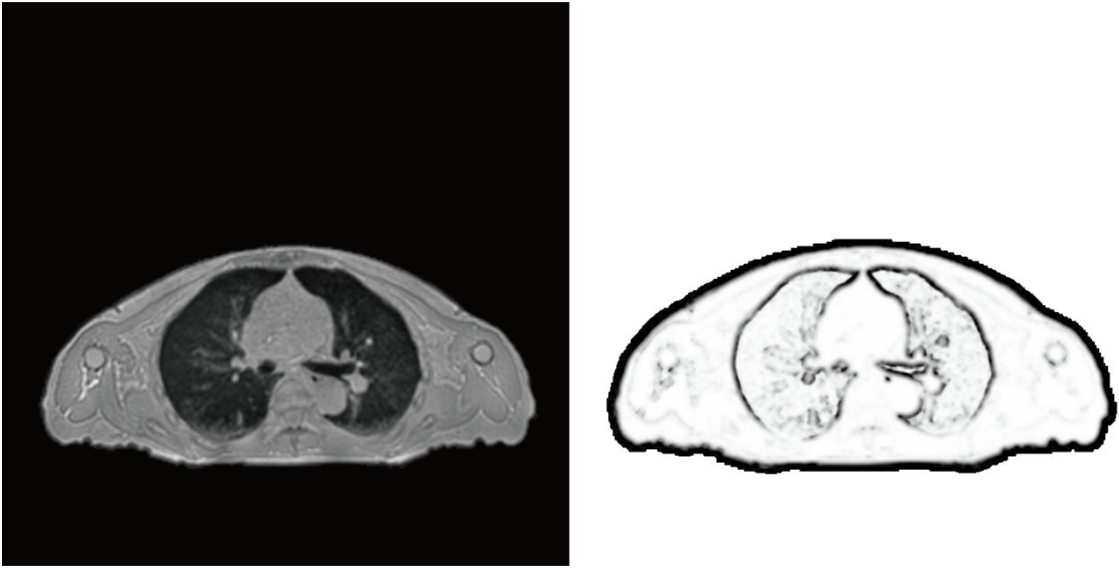

**Figure 2** MRI and the corresponding MIND image.

*2.3.1 MIND loss*

By calculating the MIND on two different images and taking the difference between them, MIND can be used as a loss that adds constraints to the change in position between them. For CycleGAN and U-GAT-IT, the difference between the MIND of the image before and after conversion is used as the loss. The MIND loss is represented by Equation (14). In the equation, $I_{MIND}(CT, r)$ is the result of adapting MIND to a CT image pixel by pixel.

$$\begin{aligned} & L_{MIND}(G_{MRI \to CT}, G_{CT \to MRI}, I_{MRI}, I_{CT}) \\ & = \mathbb{E}_{ct \sim p_{(data)(I_{CT})}}[\| I_{MIND}(G_{CT \to MRI}(ct), r) - I_{MIND}(ct, r) \|_1] \\ & + \mathbb{E}_{mri \sim p_{(data)(I_{MRI})}}[\| I_{MIND}(G_{MRI \to CT}(mri), r) - I_{MIND}(mri, r) \|_1] \end{aligned} \tag{14}$$

By incorporating $L_{MIND}$ into the loss of CycleGAN and U-GAT-IT, constraints can be added to the change in structure can be added. The loss of CycleGAN and U-GAT-IT with the addition of MIND is expressed by Equations (15) and (16):

$$L_{CycleGAN+MIND}(G_{MRI \to CT}, G_{CT \to MRI}, D_{MRI}, D_{CT}, I_{MRI}, I_{CT})$$

$$= L_{CYCLEGAN} + \lambda_{MIND} L_{MIND}(G_{MRI \to CT}, G_{CT \to MRI}, I_{MRI}, I_{CT}) \quad (15)$$

$$L_{U-GAT-IT+MIND}(G_{MRI \to CT}, G_{CT \to MRI}, D_{MRI}, D_{CT}, I_{MRI}, I_{CT}, \eta_{MRI}, \eta_{CT}, \eta D_{MRI}, \eta D_{CT})$$

$$= L_{U-GAT-IT} + \lambda_{MIND} L_{MIND}(G_{MRI \to CT}, G_{CT \to MRI}, I_{MRI}, I_{CT}) \quad (16)$$

where $\lambda_{MIND}$ denotes a coefficient of MIND loss. Finally, the models were trained by reducing the loss in Equations (17) and (18):

$$\arg \min_{G_{MRI \to CT}, G_{CT \to MRI}} \max_{D_{MRI}, D_{CT}} L_{CycleGAN+MIND} \quad (17)$$

$$\arg \min_{G_{MRI \to CT}, G_{CT \to MRI}, \eta_{MRI}, \eta_{CT}} \max_{D_{MRI}, D_{CT}, \eta D_{MRI}, \eta D_{CT}} L_{U-GAT-IT+MIND} \quad (18)$$

### 3. Experiments

This study conformed to the Declaration of Helsinki and the Ethical Guidelines for Medical and Health Research Involving Human Subjects in Japan (https://www.mhlw.go.jp/file/06-Seisakujouhou-10600000-Daijinkanboukouseikagakuka/0000080278.pdf). This study was approved by The Ethics Committee at Kobe University Graduate School of Medicine (Approval number : 170032) and was carried out according to the guidelines of the committee. The Ethics Committee at Kobe University Graduate School of Medicine has waived the need for an informed consent.

3.1 In-phase ZTE acquisition on PET/MRI

All PET/MRI examinations (n = 150; mean age, 65.9±13.0 years ; range 19 to 90 years) were performed on an integrated PET/MRI scanner (SIGNA PET/MR, GE Healthcare, Waukesha, WI, USA) at 3.0 T magnetic field strength. MR imaging of the thoracic bed position was performed with the ZTE sequence and was simultaneously acquired with a PET emission scan. No contrast-enhancing material was used. Free-breathing ZTE was acquired by three-dimensional (3D) centre-out radial sampling to provide an isotropic resolution of 2 mm$^3$, large field of view of 50 cm$^3$, and a minimal TE of zero with the following parameters: TR, ~1.4 ms; FA, 1°; 250,000 radial centre-out spokes; matrix size, 250×250; FOV, 50.0 cm$^3$; resolution, 2 mm$^3$; number of spokes per segment, 512; and approximate acquisition time, 5 min. To minimise fat–water chemical shift effects (i.e. destructive interference at fat–water tissue boundaries), a high imaging bandwidth of ±62.5 kHz was used. Furthermore, the imaging centre frequency was adjusted to be between fat and water resulting in clean in-phase ZTE images with uniform soft-tissue signal response and minimal fat–water interference disturbances [34, 35].

3.2 CT component of PET/CT

The CT component of PET/CT (Discovery PET/CT 690 (GE Healthcare), number of scans = 150; mean age, 64.4±13.9 years; range, 12–86 years) was utilised for training the CT data. The training data of ZTE and CT were acquired from different patients (unpaired datasets); however, ZTE and CT were performed in the same body position (arms down) on the respective scanners. CT was acquired during shallow expiratory breath-holding for attenuation correction of PET and acquisition of anatomical details with the following parameters: X-ray tube peak voltage (kVp), 120 kV; tube current, 20 mA; section thickness, 3.27 mm; reconstructed diameter, 500 mm; reconstructed convolutional kernel, soft.

3.3 Dataset splitting

Data of thirty cases (20%) were used as the validation dataset, and data of the remaining 120 cases (80%) were used as the training dataset. For each case, unpaired CT and ZTE were used, and no manual annotations were performed.

3.4 Image postprocessing

ZTE images were semi-automatically processed to remove the background signals by using a thresholding and filling-in technique on a commercially available workstation (Advantage workstation, GE Healthcare) and converted into a matrix size of 640 × 400. To correct the variations in sensitivity and normalise the images of ZTE to the median tissue value, the nonparametric N4ITK method was applied [36, 37]. CT images were also modified to remove the scanner beds on the workstation and were converted into the same matrix size. The MRI was maintained at the window width and window level stored in DICOM images, whereas the CT image was adjusted to a window width of 2000 Hounsfield Unit (HU) and a window level of 350 HU. The CT images were then scaled down to an image resolution of 256 × 256 pixels owing to GPU memory limitations.

3.5 Model training

All processing was performed using a workstation (CPU: Core i7-9800X at 3.80 GHz, RAM 64 GB, GPU: TITAN RTX) in all cases of CycleGAN, CycleGAN+MIND, U-GAT-IT, and U-GAT-IT+MIND.

*3.5.1 CycleGAN / CycleGAN+MIND*

We used a program based on the PyTorch implementation of CycleGAN [38], which was modified for DICOM images and MIND calculations. We used values of 10, 0.5, and 20 for $\lambda_1$, $\lambda_2$, and $\lambda_{MIND}$, respectively, in CycleGAN+MIND with Adam as the optimiser and a learning rate of 0.0002 up to 1000 epochs. A radiologist (4 years of experience) visually evaluated the results when the loss reached equilibrium. If no corruption of synthesised CT was confirmed for the training and validation datasets, the trained network was used for the main visual evaluation described below. Except for $\lambda_{MIND}$, the hyperparameters of CycleGAN and CycleGAN+MIND were the same.

*3.5.2 U-GAT-IT / U-GAT-IT+MIND*

We used a program based on the PyTorch implementation of U-GAT-IT [20], which was modified for DICOM images and MIND calculations. We used 100 for $\lambda_1$, 100 for $\lambda_2$, 100 for $\lambda_3$, and 5000 for $\lambda_{MIND}$ in U-GAT-IT +MIND, with Adam as the optimiser and a learning rate of 0.0001 up to 100 epochs. The results when the loss reached equilibrium and the training data were not corrupted by visual confirmation by the radiologist were used for evaluation. Except for $\lambda_{MIND}$, the hyperparameters of U-GAT-IT and U-GAT-IT+MIND were the same.

3.6 Visual evaluation

Twenty-one cases of chest ZTE unused for the training and validation datasets were prepared as the test dataset. The test dataset did not contain any CT images. The synthesised CTs were calculated using CycleGAN, CycleGAN+MIND, U-GAT-IT, and U-GAT-IT+MIND based on axial cross-sectional ZTE images of the supraclavicular fossa, central humeral head, sternoclavicular joint, aortic arch, tracheal bifurcation, and right pulmonary vein levels in each case. In this study, the main purpose was the application of PET/MRI attenuation-correction maps; therefore, it was particularly important to suppress the difference in anatomical structure during the conversion. For this purpose, four radiologists evaluated the synthesised CT visually, as shown below. Before evaluation by the four radiologists, a radiologist (15-year experiments) evaluated the synthesised CT images, and almost all of them were rated as CT-like for CycleGAN, CycleGAN+MIND, U-GAT-IT, and U-GAT-IT+MIND.

*3.6.1 Evaluation of image misalignment after conversion*

Visual evaluation was performed by four radiologists (Dr A, B, C, and D with 4, 22, 15, and 4 years of experience, respectively). The alignment between the synthesised CTs and the original images of the ZTE was visually evaluated for bone structures. When a relatively large defect, large displacement, or large deformation of the shape of the bone structures was observed, they were classified as having a severe misalignment. When a relatively small defect, small displacement, or small deformation of the shape of the bone structures was observed, they were classified as having a minor misalignment. One point was given when a total of 10 or more major misalignments were found in six images; two points when a total of five or more major misalignments were found; 3 points when a total of three or more major misalignments or 15 or more minor misalignments were found; 4 points when a total of 1 or more major misalignments or 10 or more minor misalignments were found; and 5 points for the others.

3.7 Statistics

The U-GAT-IT+MIND and other groups were compared using the Wilcoxon signed-rank test to evaluate the visual evaluation scores of the four radiologists. The Bonferroni method was used to correct multiple comparisons, and statistical significance was set at $p<0.001$.

**4. Results**

4.1 Synthesised CT

In Figures 3 and 4, the top images are MRI images, and those in the second to fifth rows are the synthesised CTs. Figures 3b and 4b show the fused images obtained from MRI and synthesised CT. The original ZTE images were synthesised in grey, and the synthesised CT image was in colour. Figures 3c and 4c show cropped images around the humerus in the original MR and fused images obtained from MR and synthesised CT. The displacement between the original ZTE images and the synthesised CT, especially in the body contour and the bone area, is improved by U-GAT-IT+MIND.

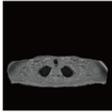

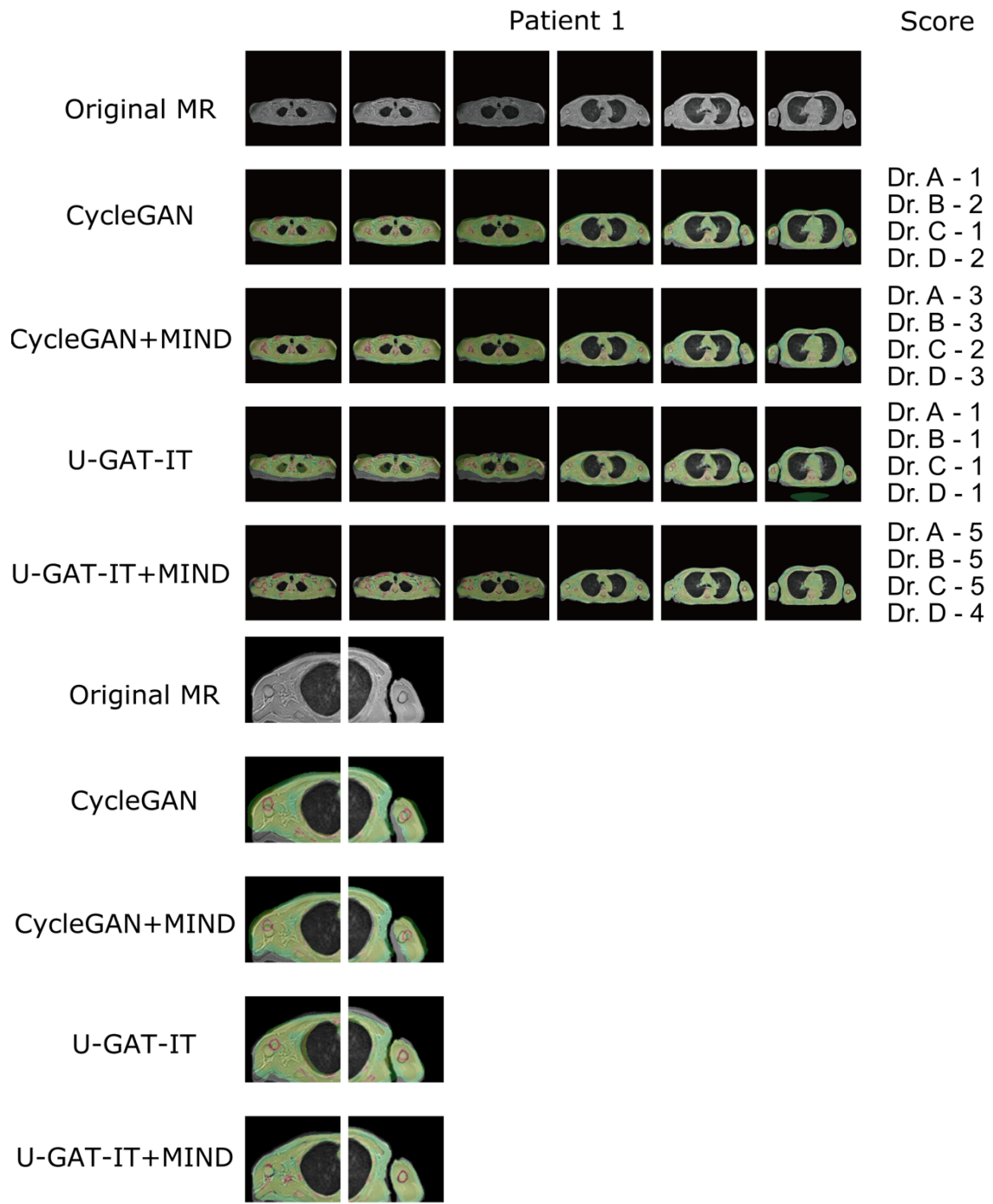

**Figure 3** Original MR and synthesised CT images along with visual evaluation scores for Patient 1.

(a) Original MR and synthesised CT images. (b) Original MR and fused images obtained from MR and synthesised CT. (c) Cropped images around the humerus in the original MR and fused images obtained from MR and synthesised CT.

Patient 2 | Score

Original MR 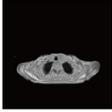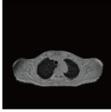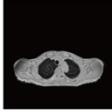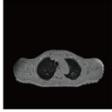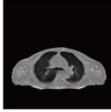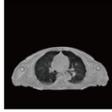

CycleGAN 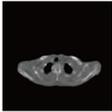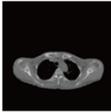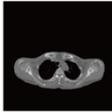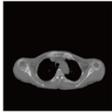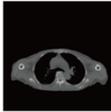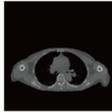
Dr. A - 3
Dr. B - 3
Dr. C - 2
Dr. D - 2

CycleGAN+MIND 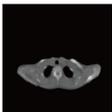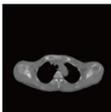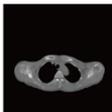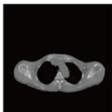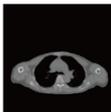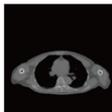
Dr. A - 3
Dr. B - 3
Dr. C - 2
Dr. D - 2

U-GAT-IT 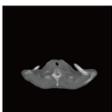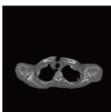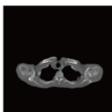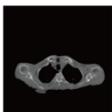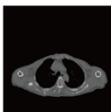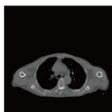
Dr. A - 3
Dr. B - 3
Dr. C - 2
Dr. D - 3

U-GAT-IT+MIND 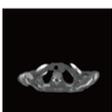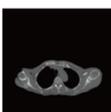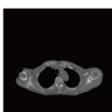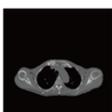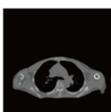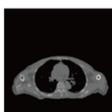
Dr. A - 5
Dr. B - 5
Dr. C - 4
Dr. D - 5

Patient 2 | Score

Original MR 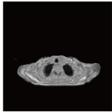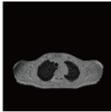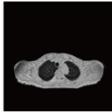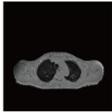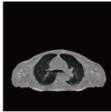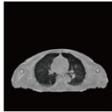

CycleGAN 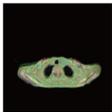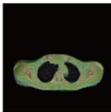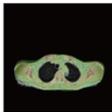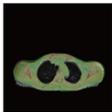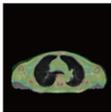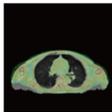
Dr. A - 3
Dr. B - 3
Dr. C - 2
Dr. D - 2

CycleGAN+MIND 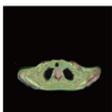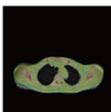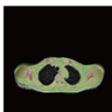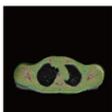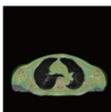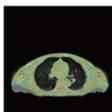
Dr. A - 3
Dr. B - 3
Dr. C - 2
Dr. D - 2

U-GAT-IT 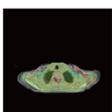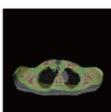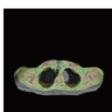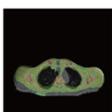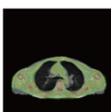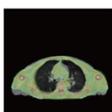
Dr. A - 3
Dr. B - 3
Dr. C - 2
Dr. D - 3

U-GAT-IT+MIND 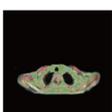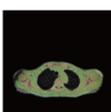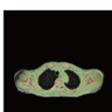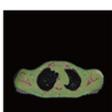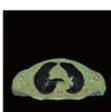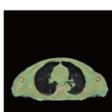
Dr. A - 5
Dr. B - 5
Dr. C - 4
Dr. D - 5

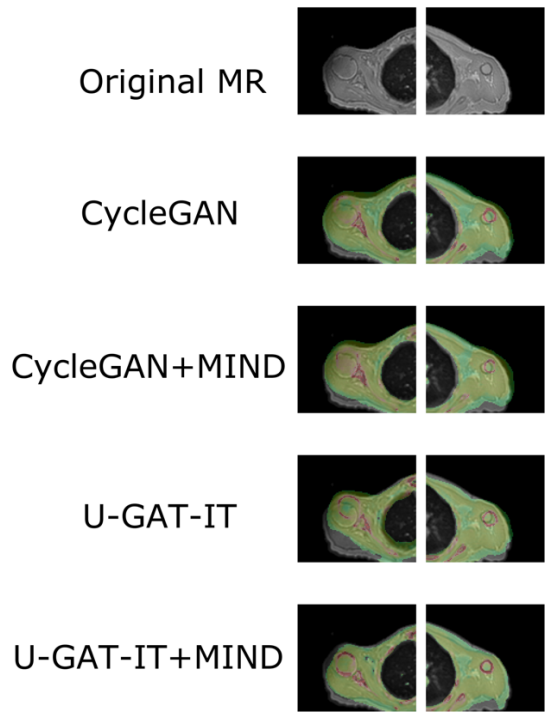

**Figure 4** Original MR and synthesised CT images along with visual evaluation scores for Patient 2.
(a) Original MR and synthesised CT images. (b) Original MR and fused images obtained from MR and synthesised CT. (c) Cropped images around the humerus in the original MR and fused images obtained from MR and synthesised CT.

Figure 5 shows the 3D VR bone images of the front and side views composited from the synthesised CT. In general, it is extremely difficult or impossible to synthesise these kinds of VR images of bone from MR images.

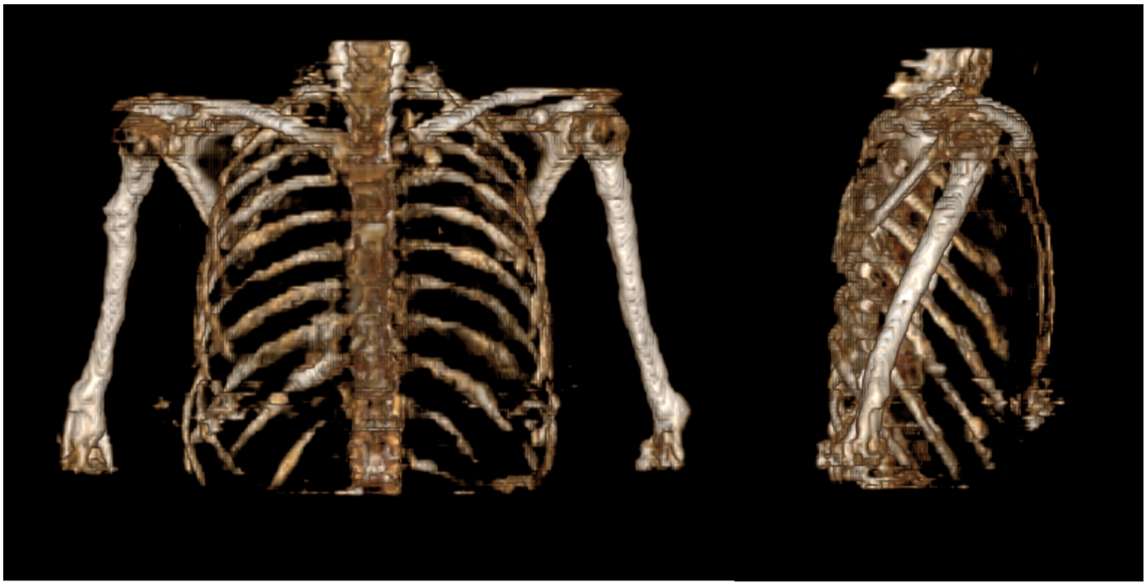

**Figure 5** Three-dimensional VR images of bones obtained via synthesised CT; in general, it is extremely difficult or almost impossible to synthesise these kinds of VR images of bone from MR images.

The 3D animation movie is attached as the Supplementary material.

The upper row of Figure 6 shows the synthesised CT based on the combination of the proposed method and conventional four-tissue segmentation, and the lower row shows the synthesised CT based on conventional four-tissue segmentation. The lower row images are clinically used for attenuation correction of PET/MRI. The upper row shows bone structures, which could not be synthesised using the conventional synthetic method (the lower row).

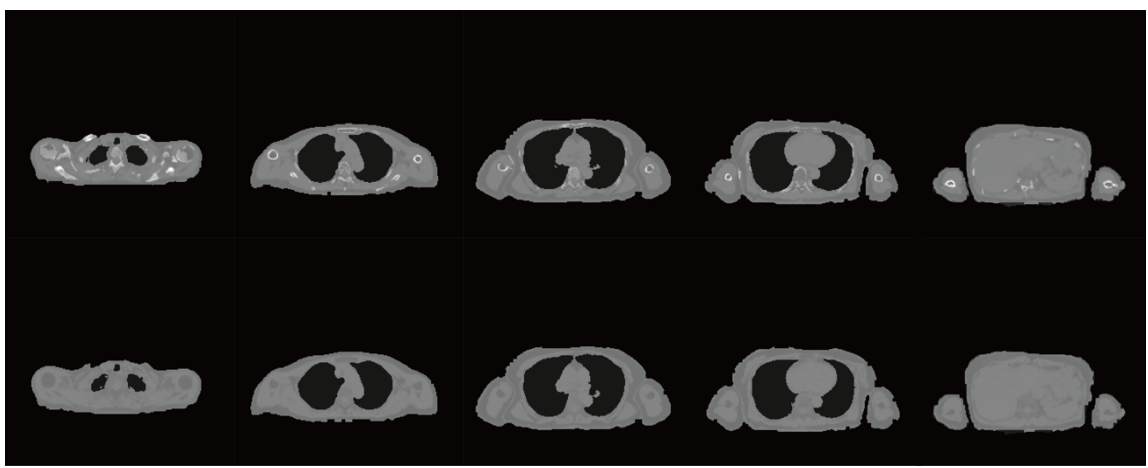

**Figure 6** Synthesised CTs using U-GAT-IT + MIND and the conventional method.

The upper row shows the combination of U-GAT-IT + MIND and conventional four-tissue segmentation. The lower row shows the synthesised CT based on the conventional four-tissue segmentation.

4.2 Visual evaluation

The results of the visual evaluation scores are summarised as follows.
- CycleGAN: maximum = 4, minimum = 1, and median = 2 by Dr. A and maximum = 4, minimum = 2, and median = 3 by Dr. B, maximum = 2, minimum = 1, and median = 2 by Dr. C, and maximum = 3, minimum = 2, and median = 2 by Dr. D
- CycleGAN + MIND: maximum = 4, minimum = 1, and median = 2 by Dr. A and maximum = 4, minimum = 2, and median = 3 by Dr. B, maximum = 2, minimum = 2, and median = 2 by Dr. C, and maximum = 3, minimum = 1, and median = 2 by Dr. D

- U-GAT-IT: maximum = 3, minimum = 1, and median = 1 by Dr. A and maximum = 4, minimum = 1, and median = 3 by Dr. B, maximum = 2, minimum = 1, and median = 1 by Dr. C, and maximum = 3, minimum = 1, and median = 2 by Dr. D
- U-GAT-IT + MIND: maximum = 5, minimum = 3, and median = 5 by Dr. A and maximum = 5, minimum = 3, and median = 5 by Dr. B, maximum = 5, minimum = 4, and median = 2 by Dr. C, and maximum = 5, minimum = 4, and median = 5 by Dr. D

Figures 7 – 10 show the distribution of visual evaluation scores by radiologists for CycleGAN, CycleGAN+MIND, U-GAT-IT, and U-GAT-IT+MIND. The boxplot of the scores of the four groups is shown in Figure 11, and the pair-plot is shown in Sup. Figures 1 – 4. The results of the Wilcoxon signed-rank test show that U-GAT-IT+MIND was significantly better than CycleGAN, CycleGAN+MIND, and U-GAT-IT. (Dr. A, $p < 0.00001$, $p < 0.00001$, $p < 0.00001$; Dr. B, $p = 0.00008$, $p = 0.00008$, $p < 0.00001$; Dr. C, $p < 0.00001$, $p = 0.00008$, $p = 0.00008$; Dr. D, $p < 0.00001$, $p < 0.00001$, $p < 0.00001$, respectively). The square indicates the median score. U-GAT-IT+MIND shows a tendency of higher scores from all radiologists.

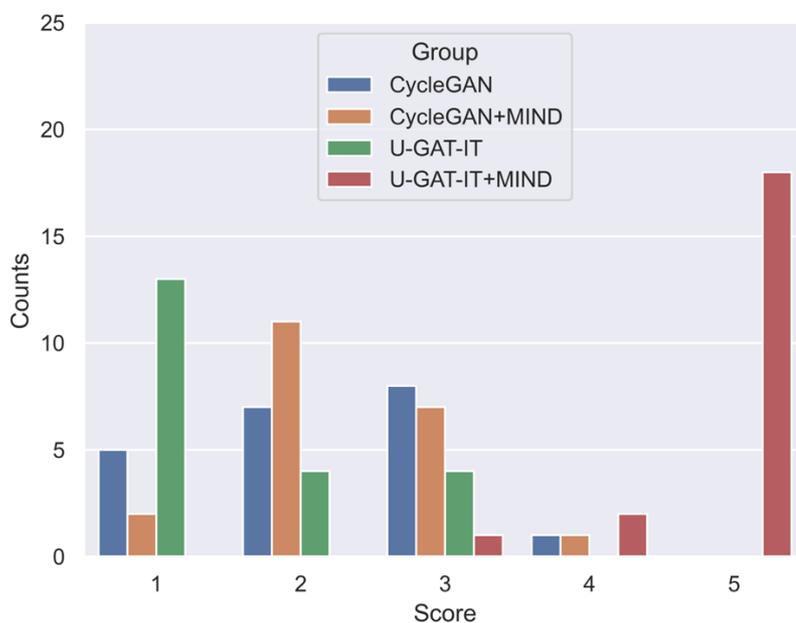

**Figure 7** Distribution of evaluation scores for four groups by Dr. A.

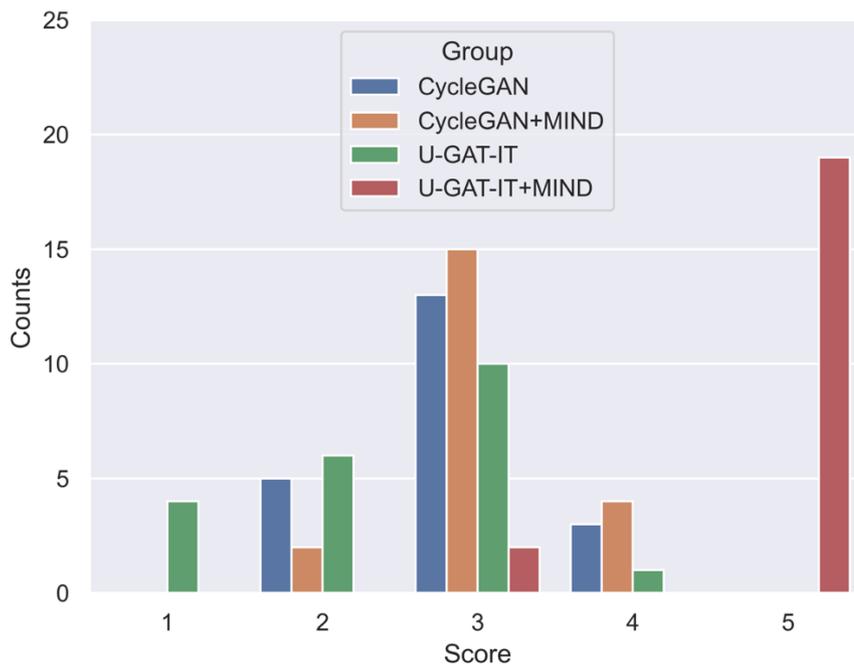

**Figure 8** Distribution of evaluation scores for four groups by Dr. B.

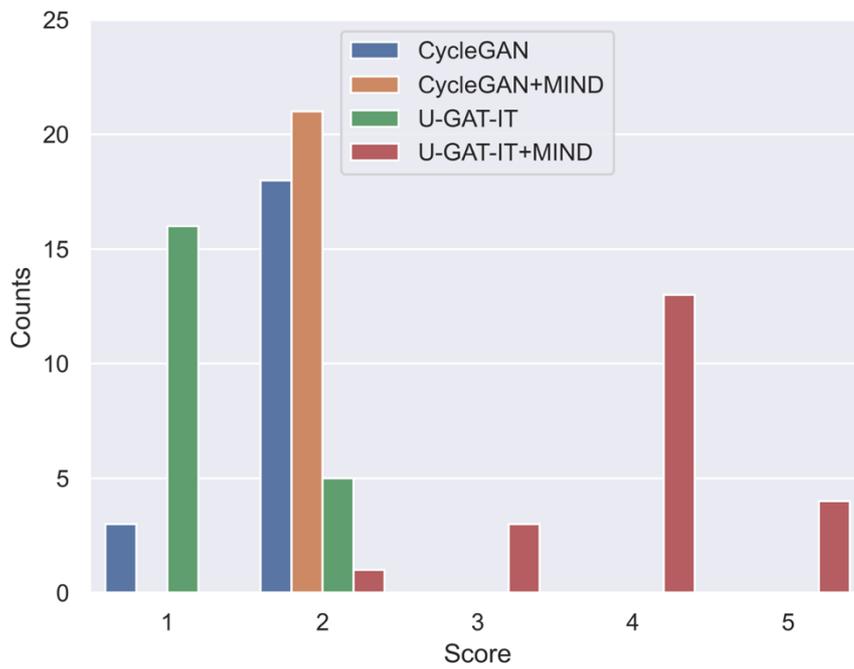

**Figure 9** Distribution of evaluation scores for four groups by Dr. C.

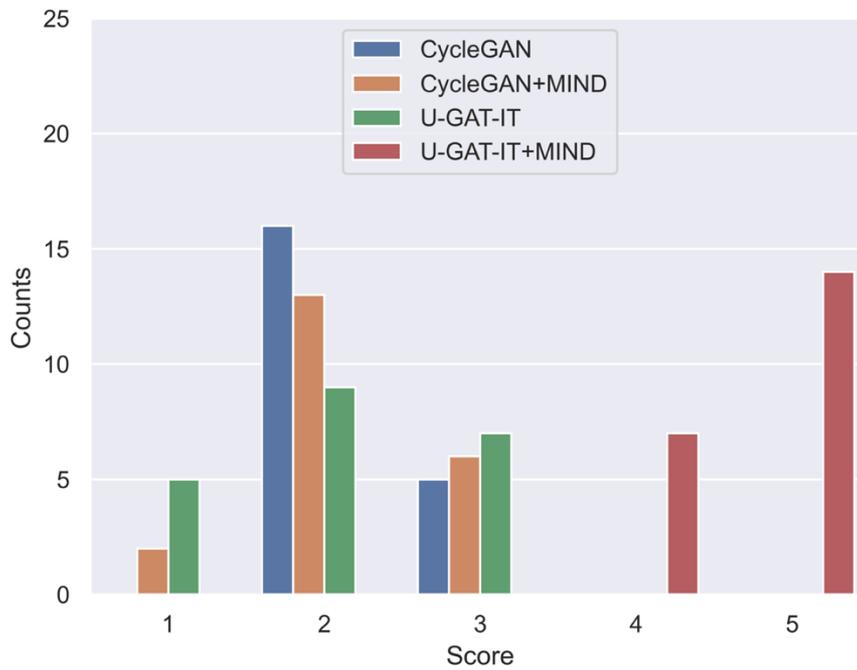

**Figure 10** Distribution of evaluation scores for four groups by Dr. D.

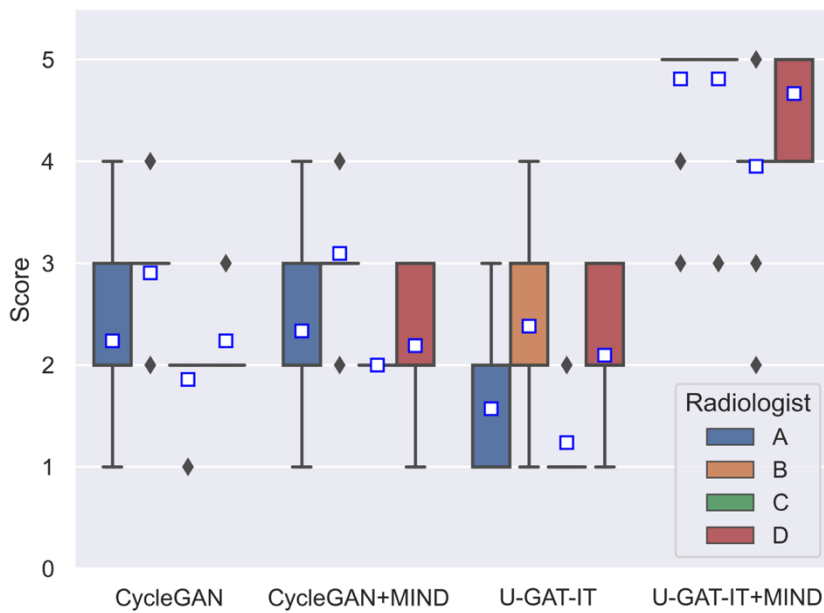

**Figure 11** Boxplot of visual evaluation scores (Note: small squares indicate the median score). As can be seen, the U-GAT-IT + MIND approach demonstrates higher scores by all Drs.]

4.3 CycleGAN+MIND with high coefficient of MIND loss

The larger the coefficient of MIND loss in CycleGAN+MIND, the more collapsed the synthesised CT became, thus distorting its contrast. Figure 12 shows the synthesised CTs from CycleGAN+MIND with a high coefficient of MIND loss ($\lambda_{MIND}$ = 60), which were apparently different from those of a normal CT. Thus, the coefficients of MIND loss of CycleGAN + MIND could not be increased to the same value as the coefficients of MIND loss of U-GAT-IT + MIND.

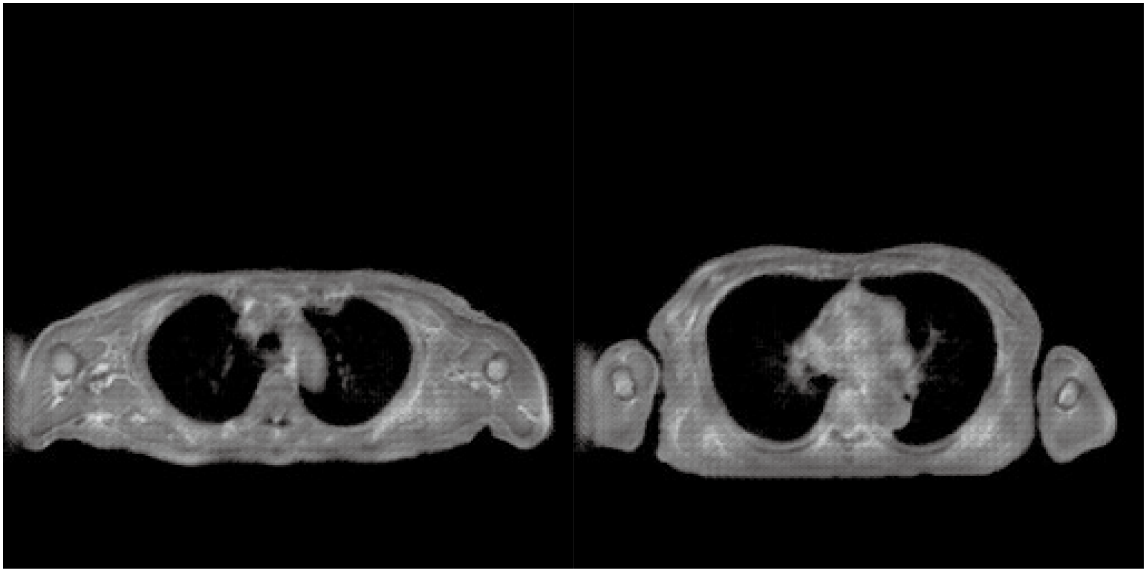

**Figure 12** CT images synthesised using CycleGAN + MIND with high MIND loss coefficient.

5. Discussion

The combination of U-GAT-IT and MIND can help in image conversion between MRI and CT images with smaller misregistration compared to conventional unpaired image transfer (CycleGAN) using unpaired datasets. The generation of paired datasets for training is simple for the head, neck, and pelvis regions because changes in body position and deformation of organs between different scans are anatomically small, which allows simple non-rigid registrations to adjust the paired data in the hand, neck, or pelvis regions. However, in the chest region, manual annotations or registrations are required to generate the paired datasets, which makes the process extremely time-consuming; furthermore, such models lack robustness because of the anatomically complicated structures of the chest and significant changes and deformation of the images between scans due to different respiratory motions, different scanner-bed shapes, and different body positions, which were the strong motivators to develop an unsupervised method for image conversion with unpaired datasets in this study. Because the annotation of bone structures is not practically possible for the chest region, it is difficult to perform quantitative evaluation; hence, visual evaluations were performed.

In this work, we also tried the combination of CycleGAN and MIND; however, the generated images were apparently different in contrast to a normal CT when the coefficient of loss by MIND ($\lambda_{MIND}$) was increased to the same range as that used for U-GAT-IT+MIND. the CAM loss introduced in U-GAT-IT prevents inconsistencies caused by the increase in MIND loss. When the coefficients of CAM loss were reduced without changing the other coefficients, the generated images seemed not to be CT-like in contrast, suggesting that the effect of CAM loss on the conversion between images was important.

There are some limitations to our study. First, we did not evaluate the effect of the synthesised CT on PET accumulation (e.g., changes in SUV) in this study. Further studies are required to confirm this hypothesis. Second, our study was conducted with a single PET/MRI scanner at a single institution, and external validation was not performed. Because the number of distributed PET/MRI scanners is limited, external validation with multiple PET/MRI scanners is difficult. Because both CycleGAN and U-GAT-IT are image conversion techniques based on unsupervised learning, the effect of overfitting is expected to be low. Fourth, it was difficult to obtain the ground truth after conversion due to the different positions and breathing conditions during PET/MRI and CT imaging, and therefore it was difficult to quantitatively evaluate the effect of the synthesised CT on PET. Further studies are required to confirm this hypothesis.

## 6. Conclusions

The combination of U-GAT-IT and MIND was effective in preventing anatomical inconsistencies between ZTE and synthesised CT and enabled the generation of clinically acceptable synthesised CT images. Our method also enables inter-modality image conversion in the chest region, which has been challenging to accomplish up until now without using human annotations.

**Data availability**

Japanese privacy protection laws and related regulations prohibit us from revealing any health-related private information such as medical images to the public without written consent, although the laws and related regulations allow researchers to use such health-related private information for research purpose under opt-out consent. We utilized the images under acceptance of the ethical committee of Kobe University Hospital under opt-out consent. It is almost impossible to take written consent to open the data to the public from all patients. For data access of our de-identified health-related private information, please contact Kobe University Hospital. The request for data access can be sent to the following e-mail addresses : hidetoshi.matsuo@bear.kobe-u.ac.jp. The other data are available from the corresponding author.

**References**

1. Zhu, J. -Y., Park, T., Isola, P. & Efros, A. A. Unpaired image-to-image translation using cycle-consistent adversarial networks. *Proceedings of the IEEE international Conference on Computer Vision*, 2223-2232 (2017).
2. Goodfellow, I. J. *et al.* Generative adversarial nets. *Adv. Neural Inf. Process. Syst.* **27** (2014).
3. Wollenweber, S. D. et al. Comparison of 4-class and continuous fat/water methods for whole-body, MR-based PET attenuation correction. *IEEE Trans. Nucl. Sci.* **60,** 3391–3398 (2013). https://doi.org/10.1109/TNS.2013.2278759
4. Gibiino, F., Sacolick, L., Menini, A., Landini, L. & Wiesinger, F. Free-breathing, zero-TE MR lung imaging. *Magn. Reson. Mater. Phys. Biol. Med.* **28,** 207–215 (2015). https://doi.org/10.1007/s10334-014-0459-y
5. Grodzki, D. M., Jakob, P. M. & Heismann, B. Correcting slice selectivity in hard pulse sequences. *J. Magn. Reson.* **214,** 61–67 (2012). https://doi.org/10.1016/j.jmr.2011.10.005
6. Madio, D. P. & Lowe, I. J. Ultra-fast imaging using low flip angles and fids. **Magn. Reason. Med. 34,** 525–529 (1995). https://doi.org/10.1002/mrm.1910340407
7. Weiger, M., Pruessmann, K. P. & Hennel, F. MRI with zero echo time: Hard versus sweep pulse excitation. *Magn. Reason. Med.* **66,** 379–389 (2011). https://doi.org/10.1002/mrm.22799
8. Wiesinger, F. et al. Zero TE MR bone imaging in the head. *Magn. Reson. Med.* **75,** 107–114 (2016). https://doi.org/10.1002/mrm.25545
9. Wu, Y. et al. Density of organic matrix of native mineralized bone measured by water- and fat-suppressed proton projection MRI. *Magn. Reason. Med.* **50,** 59–68 (2003). https://doi.org/10.1002/mrm.10512
10. Aasheim, L. B. et al. PET/MR brain imaging: Evaluation of clinical UTE-based attenuation correction. *Eur. J. Nucl. Med. Mol. Imaging.* **42,** 1439–1446 (2015). https://doi.org/10.1007/s00259-015-3060-3
11. Delso, G. et al. Clinical evaluation of zero-echo-time MR imaging for the segmentation of the skull. *J. Nucl. Med.* **56,** 417–422 (2015). https://doi.org/10.2967/jnumed.114.149997
12. Sekine, T. et al. Clinical evaluation of zero-echo-time attenuation correction for brain 18F-FDG PET/MRI: Comparison with atlas attenuation correction. *J. Nucl. Med.* **57,** 1927–1932 (2016). https://doi.org/10.2967/jnumed.116.175398

13. Sgard, B. et al. ZTE MR-based attenuation correction in brain FDG-PET/MR: Performance in patients with cognitive impairment. *Eur. Radiol.* **30,** 1770–1779 (2020). https://doi.org/10.1007/s00330-019-06514-z
14. Bradshaw, T. J., Zhao, G., Jang, H., Liu, F., & McMillan, A. B. Feasibility of deep learning-based PET/MR attenuation correction in the pelvis using only diagnostic MR images. *Tomography* **4,** 138–147 (2018). https://doi.org/10.18383/j.tom.2018.00016
15. Leynes, A. P. et al. Zero-echo-time and dixon deep pseudo-CT (ZeDD CT): Direct generation of pseudo-CT images for pelvic PET/MRI attenuation correction using deep convolutional neural networks with multiparametric MRI. *J. Nucl. Med.* **59,** 852–858 (2018). https://doi.org/10.2967/jnumed.117.198051
16. Torrado-Carvajal, A. et al. Dixon-vibe deep learning (DIVIDE) pseudo-CT synthesis for pelvis PET/MR attenuation correction. *J. Nucl. Med.* **60,** 429–435 (2019). https://doi.org/10.2967/jnumed.118.209288
17. Liang, X. et al. Generating synthesized computed tomography (CT) from cone-beam computed tomography (CBCT) using CycleGAN for adaptive radiation therapy. *Phys. Med. Biol.* **64,** 125002 (2019). https://doi.org/10.1088/1361-6560/ab22f9
18. Lei, Y. et al. MRI-only based synthetic CT generation using dense cycle consistent generative adversarial networks. *Med. Phys.* **46,** 3565–3581 (2019). https://doi.org/10.1002/mp.13617
19. Tang, Y., Tang, Y., Xiao, J., & Summers, R. M. XLSor: A robust and accurate lung segmentor on chest X-Rays using criss-cross attention and customized radiorealistic abnormalities generation. *International Conference on Medical Imaging with Deep Learning,* 457-467 (2019).
20. Kim, J., Kim, M., Kang, H., & Lee, K. U-GAT-IT: Unsupervised generative attentional networks with adaptive layer-instance normalization for image-to-image translation. *arXiv preprint arXiv:1907.10830* (2019).
21. Heinrich, M. P. et al. AMIND: Modality independent neighbourhood descriptor for multi-modal deformable registration. *Med. Image Anal.* **16,** 1423–1435 (2012). https://doi.org/10.1016/j.media.2012.05.008
22. Lei, Y., Jeong, J. J. & Wang, T. MRI-based pseudo CT synthesis using anatomical signature and alternating random forest with iterative refinement model. *J. Med. Imaging* **5,** 1 (2018). https://doi.org/10.1117/1.jmi.5.4.043504
23. Torrado-Carvajal, A. et al. Fast patch-based pseudo-CT synthesis from T1-weighted MR images for PET/MR attenuation correction in brain studies. *J. Nucl. Med.* **57,** 136–143 (2016). https://doi.org/10.2967/jnumed.115.156299

24. Liang, X. et al. Generating synthesized computed tomography (CT) from cone-beam computed tomography (CBCT) using CycleGAN for adaptive radiation therapy. Phys. Med. Biol. 64, 125002 (2019).
25. Sandfort, V., Yan, K., Pickhardt, P. J. & Summers, R. M. Data augmentation using generative adversarial networks (CycleGAN) to improve generalizability in CT segmentation tasks. *Sci. Rep.* **9,** 16884 (2019). https://doi.org/10.1038/s41598-019-52737-x
26. Tmenova, O., Martin, R. & Duong, L. CycleGAN for style transfer in X-ray angiography. *Int. J. Comput. Assist. Rad. Surg.* **14,** 1785–1794 (2019). https://doi.org/10.1007/s11548-019-02022-z
27. Abramian, D. & Eklund, A. Generating fMRI volumes from T1-weighted volumes using 3D CycleGAN. *arXiv preprint arXiv:1907.08533* (2019).
28. Cai, J., Zhang, Z., Cui, L., Zheng, Y. & Yang, L. Towards cross-modal organ translation and segmentation: A cycle- and shape-consistent generative adversarial network. *Med. Image Anal.* **52,** 174–184. (2019). https://doi.org/10.1016/j.media.2018.12.002
29. Pan, Y. et al. Synthesizing missing PET from MRI with cycle-consistent generative adversarial networks for Alzheimer's disease diagnosis. *Lecture Notes in Computer Science (Including Subseries Lecture Notes in Artificial Intelligence and Lecture Notes in Bioinformatics). Springer Verlag,* 455–463 (2018). https://doi.org/10.1007/978-3-030-00931-1_52
30. Wang, C., Macnaught, G., Papanastasiou, G., MacGillivray, T. & Newby, D. Unsupervised learning for cross-domain medical image synthesis using deformation invariant cycle consistency networks. *Lecture Notes in Computer Science (Including Subseries Lecture Notes in Artificial Intelligence and Lecture Notes in Bioinformatics). Springer Verlag,* 52–60 (2018). https://doi.org/10.1007/978-3-030-00536-8_6
31. Yang, H. et al. Unpaired brain MR-to-CT synthesis using a structure-constrained CycleGAN. *Lecture Notes in Computer Science (Including Subseries Lecture Notes in Artificial Intelligence and Lecture Notes in Bioinformatics). Springer Verlag,* 174–182. (2018). https://doi.org/10.1007/978-3-030-00889-5_20
32. Hiasa, Y. et al. Cross-modality image synthesis from unpaired data using CycleGAN: Effects of gradient consistency loss and training data size. *Lecture Notes in Computer Science (Including Subseries Lecture Notes in Artificial Intelligence and Lecture Notes in Bioinformatics). Springer Verlag,* 31–41. (2018). https://doi.org/10.1007/978-3-030-00536-8_4

33. Harms, J. et al. Paired cycle-GAN-based image correction for quantitative cone-beam computed tomography. *Med. Phys.* **46,** 3998–4009 (2019). https://doi.org/10.1002/mp.13656
34. Brodsky, E. K., Holmes, J. H., Yu, H. & Reeder, S. B. Generalized K-space decomposition with chemical shift correction for non-Cartesian water-fat imaging. *Magn. Reason. Med.* **59,** 1151–1164 (2008). https://doi.org/10.1002/mrm.21580
35. Engström, M., McKinnon, G., Cozzini, C. & Wiesinger, F. In-phase zero TE musculoskeletal imaging. *Magn. Reason. Med.* **83,** 195–202 (2020). https://doi.org/10.1002/mrm.27928
36. Sled, J. G., Zijdenbos, A. P. & Evans, A. C. A nonparametric method for automatic correction of intensity nonuniformity in MRI data. *IEEE Trans. Med. Imaging.* **17,** 87–97. (1998). https://doi.org/10.1109/42.668698
37. Tustison, N. J. et al. N4ITK: Improved N3 bias correction. *IEEE Trans. Med. Imaging.* **29,** 1310–1320. (2010). https://doi.org/10.1109/TMI.2010.2046908
38. Zhu, J.-Y., Park, T., Isola, P. & Efros, A. A. Unpaired Image-to-Image Translation using Cycle-Consistent Adversarial Networks. (2017).

**Acknowledgement**

This work was supported by JSPS KAKENHI (Grant Number JP20K16758 and JP19H03599).

**Author contributions**

Conceptualization: HM, MNi and MNo.
Data curation: MNo.
Formal analysis: HM and MNi.
Funding acquisition: FZ and MNo.
Investigation: HM, MNi and MNo.
Methodology: HM ,NMi, MNo, TK, SK and FW.
Project administration: MNi, MNo and AKK.
Resources: HM.
Supervision: TM.
Validation: HM.
Visualization: HM.
Writing – original draft: HM, and MNi.
Writing – review & editing: MNo, FZ, TK, SK, FW, AKK and TM.
All authors approved the manuscript to be published and agree to be accountable for all aspects of the work in ensuring that questions related to the accuracy or integrity of any part of the work are appropriately investigated and resolved.

**Competing interests**

TK, SK and FW are employees of GE Healthcare.
Other authors have no competing financial interests.

**Supplementary material**

An animation of 3D VR images composited from the synthesised CT is submitted as "Supplementary Material" along with this manuscript.

The pair plots of the visual evaluation scores by radiologists for CycleGAN. CycleGAN + MIND, U-GAT-IT, and U-GAT-IT + MIND.

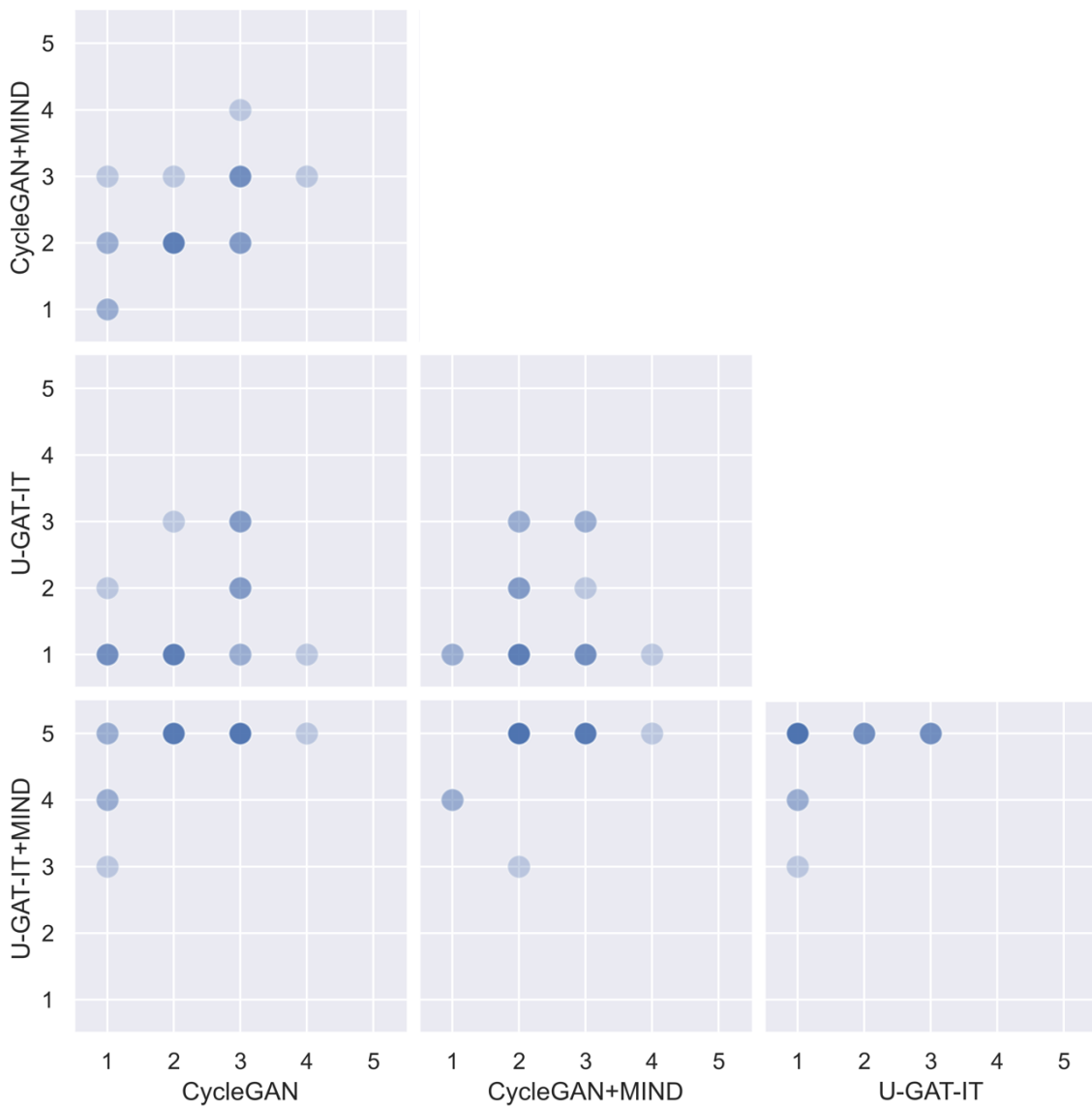

**Sup. Figure 1** Pair-plot of visual evaluation scores (Dr. A) (dark circles indicate high frequencies).

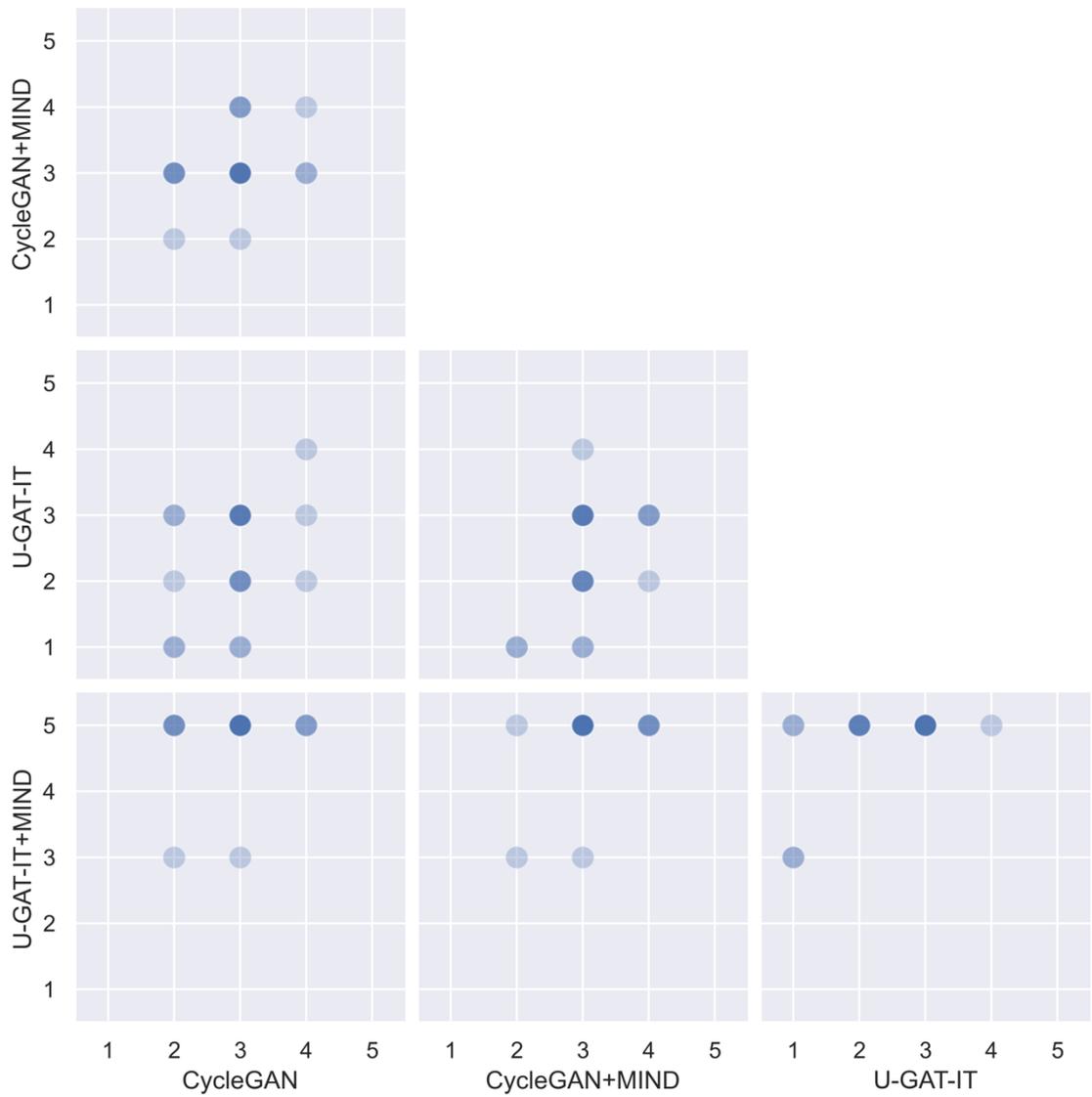

**Sup. Figure 2** Pair-plot of visual evaluation scores (Dr. B) (dark circles indicate high frequencies).

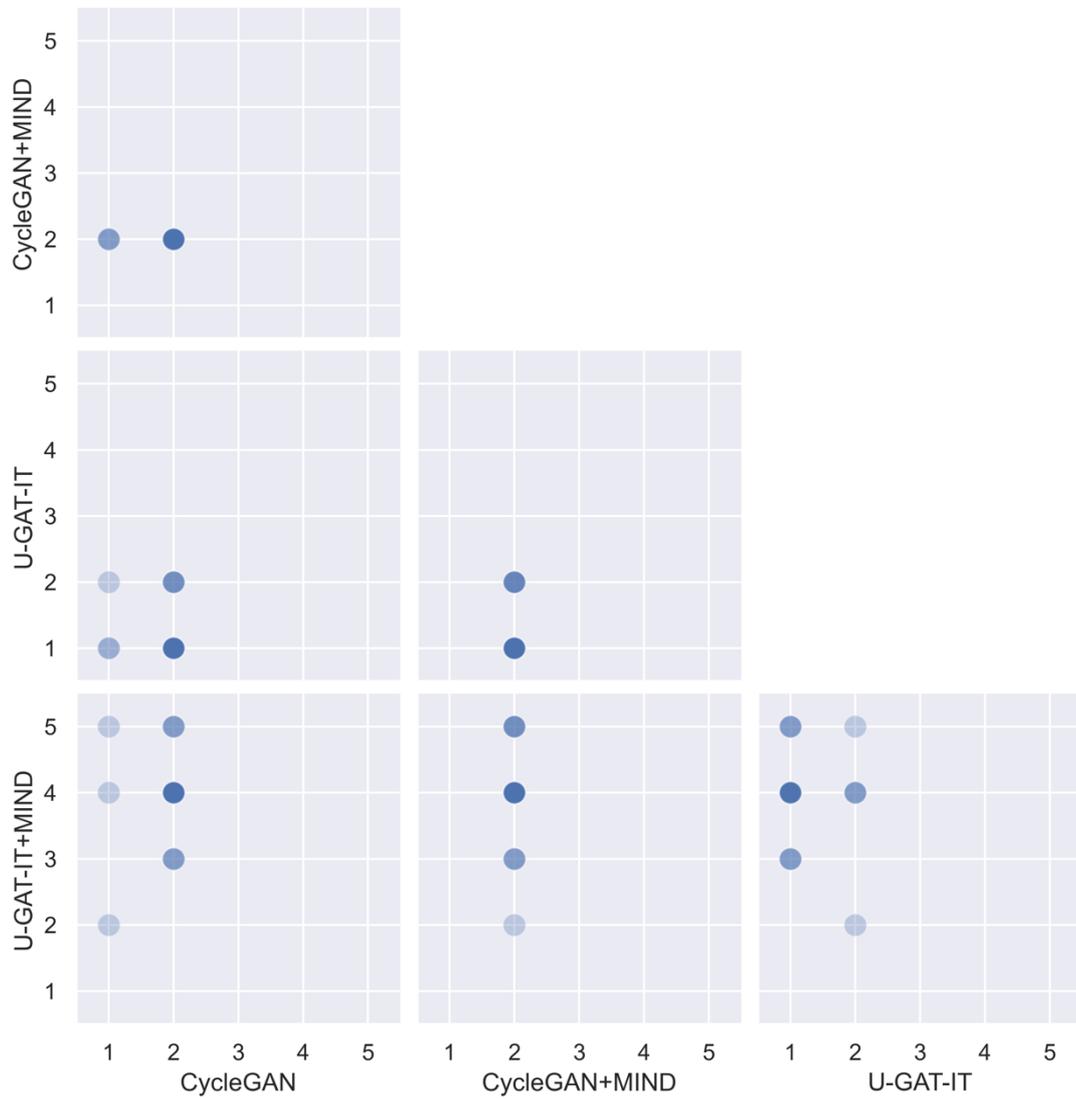

**Sup. Figure 3** Pair-plot of visual evaluation scores (Dr. C) (dark circles indicate high frequencies).

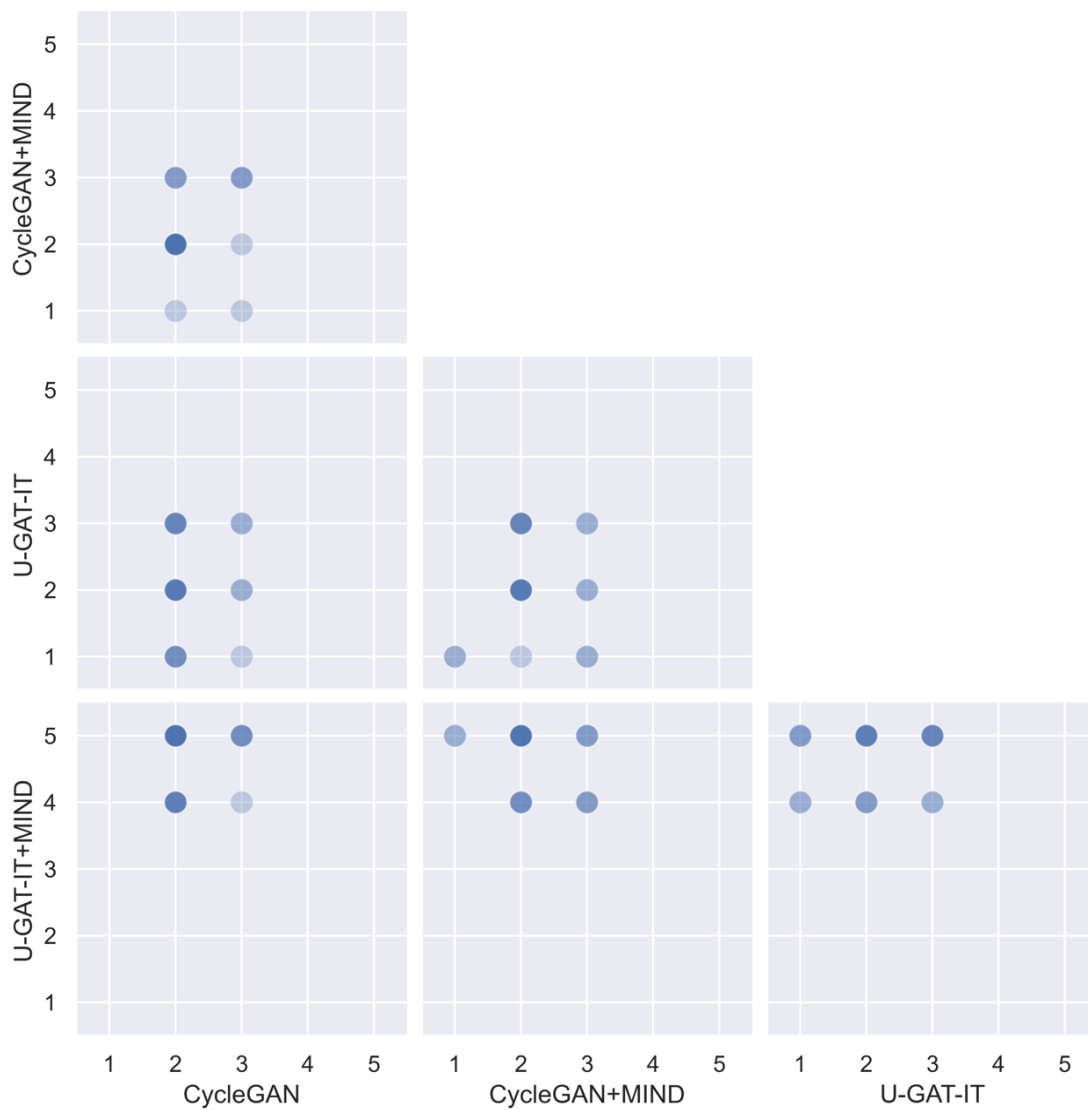

**Sup. Figure 4** Pair-plot of visual evaluation scores (Dr. D) (dark circles indicate high frequencies).